\documentclass[lettersize,journal]{IEEEtran}
\usepackage{amsmath,amsfonts}
\usepackage{bbding}
\usepackage{multirow}
\usepackage{tabu}
\usepackage{bm}
\usepackage{algorithmic}
\usepackage{algorithm}
\usepackage{array}
\usepackage[caption=false,font=normalsize,labelfont=sf,textfont=sf]{subfig}
\usepackage{textcomp}
\usepackage{stfloats}
\usepackage{url}
\usepackage{verbatim}
\usepackage{graphicx}
\usepackage{cite}
\usepackage{color}
\usepackage{booktabs}
\usepackage{verbatim}

\usepackage{makecell}

\def\etal{\emph{et al.}}
\def\ie{\emph{i.e.},}

\begin{document}

\title{Cross-Modality Proposal-guided Feature Mining for Unregistered RGB-Thermal Pedestrian Detection}

\author{Chao~Tian,
        Zikun~Zhou,
        Yuqing~Huang,
        Gaojun~Li,
        and Zhenyu~He*,~\IEEEmembership{Senior Member, IEEE}
\thanks{
    *Corresponding author.

    This research is supported by the National Natural Science Foundation of China (No.62172126), and the Shenzhen Research Council (No.JCYJ20210324120202006), and The Major Key Project of PCL (PCL2021A03-1).
    
    Chao Tian and Gaojun Li are with the School of Computer Science and Technology, Harbin Institute of Technology, Shenzhen, China (e-mail: tianchao@stu.hit.edu.cn; lee\_gaojun@163.com)
    
    Zikun Zhou is with Peng Cheng Laboratory, Shenzhen, China (e-mail: zhouzikunhit@gmail.com)

    Yuqing Huang and Zhenyu He are with the School of Computer Science and Technology, Harbin Institute of Technology, Shenzhen, China and Peng Cheng Laboratory, Shenzhen, China (e-mail: 22b951033@stu.hit.edu.cn; zhenyuhe@hit.edu.cn).
    
    Chao Tian and Zikun Zhou contributed equally to this work.
    }
}



\maketitle

\begin{abstract}
RGB-Thermal (RGB-T) pedestrian detection aims to locate the pedestrians in RGB-T image pairs to exploit the complementation between the two modalities for improving detection robustness in extreme conditions. Most existing algorithms assume that the RGB-T image pairs are well registered, while in the real world they are not aligned ideally due to parallax or different field-of-view of the cameras. The pedestrians in misaligned image pairs may locate at different positions in two images, which results in two challenges: 1) how to achieve inter-modality complementation using spatially misaligned RGB-T pedestrian patches, and 2) how to recognize the unpaired pedestrians at the boundary. To deal with these issues, we propose a new paradigm for unregistered RGB-T pedestrian detection, which predicts two separate pedestrian locations in the RGB and thermal images, respectively. Specifically, we propose a cross-modality proposal-guided feature mining (CPFM) mechanism to extract the two precise fusion features for representing the pedestrian in the two modalities, even if the RGB-T image pair is unaligned. It enables us to effectively exploit the complementation between the two modalities. With the CPFM mechanism, we build a two-stream dense detector; it predicts the two pedestrian locations in the two modalities based on the corresponding fusion feature mined by the CPFM mechanism. Besides, we design a data augmentation method, named Homography, to simulate the discrepancy in scales and views between images. We also investigate two non-maximum suppression (NMS) methods for post-processing. Favorable experimental results demonstrate the effectiveness and robustness of our method in dealing with unregistered pedestrians with different shifts.

\end{abstract}


\section{Introduction}\label{sec1}
\IEEEPARstart{P}{edestrian} detection aims to locate all pedestrians in the view~\cite{li2017scale, wang2018pedestrian, dollar2011pedestrian}. It is an important research topic in computer vision with a wide range of applications, including security surveillance~\cite{fuhr2015camera, bilal2016low} and autonomous driving~\cite{yang2018real, chen2021deep}. Numerous pedestrian detectors~\cite{li2017scale,wang2018pedestrian,dollar2009pedestrian, benenson2014ten,mao2017can,kieu2020task,herrmann2018cnn} have been proposed and achieved astonishing progress. However, most of them are designed for RGB pedestrian detection. These algorithms inevitably suffer from low imaging quality resulting from extreme conditions, such as low illumination and fog. Compared with the RGB images, the thermal images, which record the temperature of the scene, are insensitive to the illumination conditions. Hence, several methods resort to the thermal image to complement the RGB image to improve the detection robustness in extreme conditions ~\cite{hwang2015multispectral,zhang2021guided,IJWMIP}, which are so-called RGB-Thermal (RGB-T) detectors.

\begin{figure}[!t]
\centering
\includegraphics[width=\linewidth]{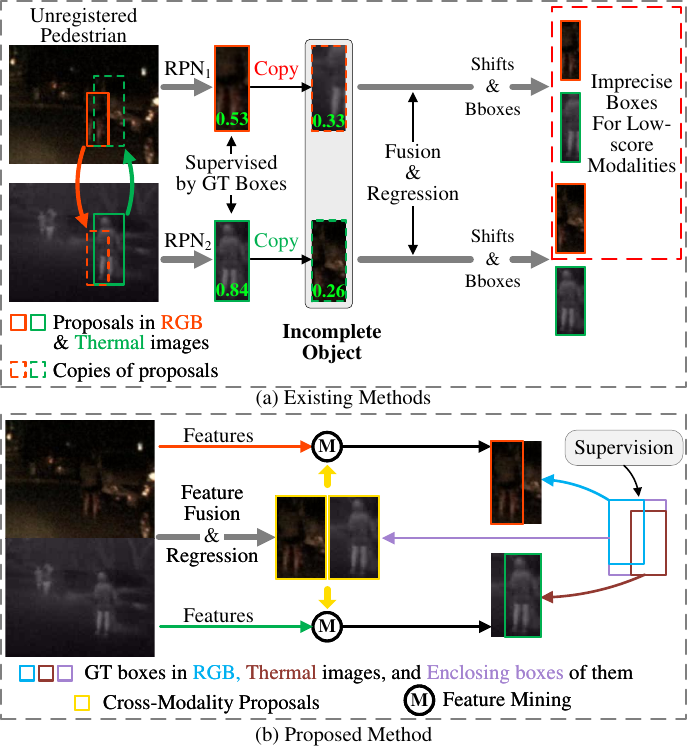}
\caption{Illustration of our motivation. (a) In existing methods, the proposals in RGB and thermal images are predicted independently, which results in incomplete objects after cross-modality cloning. The incomplete objects further lead to inaccurate cross-modality shift predictions and affect the final detection results. (b) The proposed cross-modality proposal avoids the missing of information caused by misalignment. Features characterizing the pedestrian in a certain modality are then mined adaptively to predict the precise boxes in this modality with the guidance of the cross-modality proposal.
}
\label{Fig:intro}
\end{figure}

Recent RGB-T detectors mainly focus on how to effectively fuse RGB and thermal information to make them complementary to each other. Three types of fusion strategies have been proposed: data fusion~\cite{wagner2016multispectral}, feature fusion~\cite{liu2016multispectral,zhang2021guided}, and box fusion~\cite{li2022confidence}, which achieve remarkable robustness. Nevertheless, most of them assume that the RGB-T image pair is well registered and predict the same bounding box for the RGB and thermal images. Namely, these methods overlook the unregistered issue of the RGB-T image pair caused by parallax or the different field-of-view (FOV) of the two cameras, which is quite a common yet intractable issue.

One of the challenges for unregistered RGB-T detection is how to achieve inter-modality complementation using spatially misaligned RGB and thermal pedestrian patches. Due to the misalignment issue, the background around the pedestrian is apt to be involved in the pedestrian feature fusion, which further disturbs the bounding box regression and pedestrian-background classification. Another challenge is that the pedestrian at the boundary in one image may be out of the other image due to the different FOVs, \ie~the unpaired pedestrian issue. Thus an ideal unregistered pedestrian RGB-T detector should not only predict precise bounding boxes for the paired pedestrian but also recognize the unpaired pedestrian.

AR-CNN~\cite{zhang2019weakly} is the pioneering method for unregistered RGB-T pedestrian detection. After that, several algorithms ~\cite{wanchaitanawong2021multi,ni2022modality} also take into consideration the unregistered issue for RGB-T pedestrian detection. These approaches are all based on Faster-RCNN~\cite{fasterrcnn} and deploy two independent Region Proposal Networks (RPNs) for the RGB and thermal images, respectively. In these methods, the proposal in one modality is cloned to the other modality. The features of the original proposal and the counterpart are fused together, and the fusion features are further used to regress the shift between the pedestrian locations in the RGB and thermal images, as shown in Figure~\ref{Fig:intro}(a). However, the counterpart, \ie~the cloned proposal, may not completely cover the pedestrian, as shown in Figure~\ref{Fig:intro}(a). The incomplete proposal results in part of the pedestrian information missing in the sampled RoI features, which further leads to inaccurate cross-modality shift predictions. The inaccurate shift would finally hinder us to obtain an accurate box in each modality. For example, if the RGB image is of poor quality as shown in Figure~\ref{Fig:intro}, we cannot correct the originally inaccurate RGB detection result conditioned on the thermal detection result as the predicted cross-modality shift is inaccurate. 

In this paper, we propose a new paradigm for unregistered RGB-T pedestrian detection to handle the above-mentioned two challenges. We build our algorithm based on a dense detector~\cite{redmon2016you,redmon2018yolov3,ge2021yolox}, which has shown great potential in general object detection and RGB-T pedestrian detection. To achieve effective inter-modality complementation conditioned on the unaligned RGB-T image pair, we design a Cross-Modality Feature Mining (CPFM) head. The CPFM head works in two steps: 1) first predicts the cross-modality proposal enclosing the pedestrian in two modalities based on the fusion feature, and 2) then adaptively mines the feature characterizing the pedestrian in a certain modality from the fusion feature with the guidance of the cross-modality proposal. The cross-modality proposal enables us to avoid missing the pedestrian information in a certain modality due to misalignment, as shown in Figure~\ref{Fig:intro}(b). Based on the mined features for each modality, we can predict an accurate bounding box in the corresponding modality.

To deal with the unpaired pedestrian issue, we resort to post-processing methods to remove the low-score boxes to avoid false positive results in unpaired cases. Specifically, we investigate two variants of Non-Maximum Suppression (NMS), Decoupled NMS and Pair-wise NMS, to adapt the dense detector for the unregistered RGB-T pedestrian detection task. Besides, we design a data augmentation method, named Homography, to perform homography transformation on the thermal image to simulate the unregistered RGB-T image pair with scale differences and shifts. We extensively evaluate the proposed algorithm on two popular RGB-T pedestrian detection benchmarks, KAIST~\cite{hwang2015multispectral} and CVC-14~\cite{cvc14}, which contain many unaligned pedestrians. We further simulate larger misalignment in KAIST to analyze the robustness of our method. Favorable performance on these benchmarks demonstrates the effectiveness of our detector. Our contributions can be summarized as follows:
\begin{itemize}
\item{We propose a cross-modality proposal-guided feature mining head to effectively exploit the complementation between the two modalities on the condition that the RGB-T image pair is unaligned.}

\item{We design the data augmentation and post-processing methods to adapt the dense detector to the unregistered RGB-T pedestrian detection task.}
\item{Favorable experimental results demonstrate the effectiveness and robustness of our method in dealing with unregistered pedestrians with different shifts.}
\end{itemize}

\section{Related Work}
\subsection{Object Detection}
Recently, object detection methods based on neural networks have achieved impressive progress. RCNN-like detectors~\cite{rcnn,fastrcnn,fasterrcnn,maskrcnn} are known as the pioneering works, that give CNN-based solutions to object detection. These works get the region proposals at first, then extract RoI features guided by proposals to predict the class and bounding boxes, thus known as two-stage detectors. The Feature Pyramid Network (FPN)~\cite{fpn} builds a series of features at different scales to promote multi-scale performance. 

Compared with two-stage detectors, single-stage detectors, which are trained end-to-end, get the detection results directly by dense prediction~\cite{ssd,yolo,redmon2018yolov3}. To this end, single-stage detectors must need an NMS operation at last. Single-stage detectors were once considered a kind of fast but imprecise method. Nevertheless, many modern single-stage detectors have a better performance than classic two-stage detectors, such as Faster-RCNN~\cite{fasterrcnn}. ATSS~\cite{ATSS} proves that anchor-based detectors and anchor-free ones have no difference in their potential, which is a debate in the early development of single-stage detectors. ATSS also proposes the adaptive training sample selection strategy for the first time. \cite{facalloss} decouples the regression branch and the classification branch in heads, and applies focal loss to training. Later detectors~\cite{ge2021yolox,varifocalnet,GFL} make further progress with better training sample selection strategies and loss function settings.

\begin{figure*}[!t]
\centering
\includegraphics[width=7in]{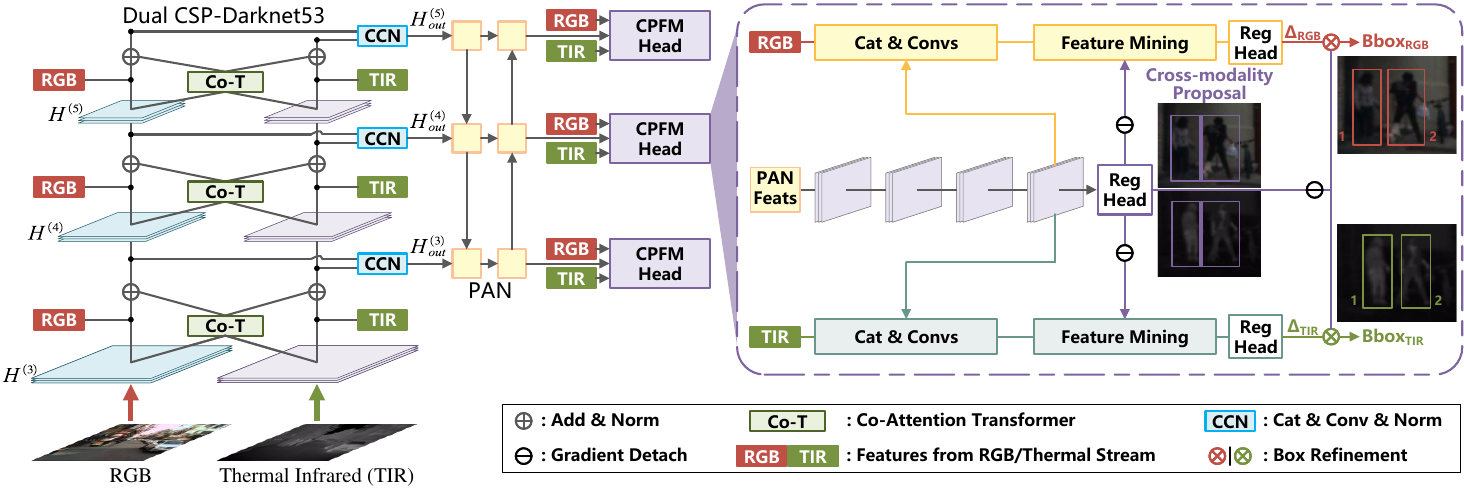}
\caption{Overall framework. The network takes dual CSP-Darknet53s as the backbone. Two branches of the backbone take RGB and thermal images respectively as input. Three Co-Attention Transformer layers are inserted at three stages between branches to strengthen the global reception, which is essential for unregistered modalities. Fusion features are fed into PAN to promote multi-scale detection ability. In the CPFM head, fusion features from PAN are used to predict cross-modality proposals. With the guidance of proposals, the feature mining mechanism is conducted on modal-enhanced features, and gets precise boxes for each modality. Decoupled heads are applied in the CPFM head, and three classification branches are not shown in the figure for concision. }
\label{Fig:framework}
\end{figure*}

\subsection{RGB-T Pedestrian Detection}
More attention has been paid to RGB-T pedestrian detection to strengthen the performance of pedestrian detection in challenging conditions. KAIST Benchmark~\cite{kaist} and CVC-14~\cite{cvc14} greatly facilitate the development of RGB-T pedestrian detection. Wagner \etal~\cite{wagner2016multispectral} propose the pioneering work of CNN-based RGB-T pedestrian detection. Liu \etal~\cite{liu2016multispectral} propose the multi-stage feature fusion between modalities for the first time, which has become a common strategy nowadays. CIAN~\cite{cian} proposes a cross-modality interactive attention module to enhance the fusion of information from different modalities, which is inserted between two streams of different modalities in the backbone. Li \etal~\cite{li2019illumination} apply an illumination-aware mechanism on RGB images to predict the credibility of RGB images according to illumination during fusion. MBNet~\cite{mbnet2020} balances the efficient information in different modalities by channel-wise differential weighting and also adopts an illumination-aware weighting strategy. Considering the weak correlation between the illumination of the whole image and the confidence of a specific proposal in RGB images, CMPD~\cite{li2022confidence} utilizes the confidence of proposals, instead of the illumination of whole images, to weight features before fusion. GAFF~\cite{zhang2021guided} conducts a dynamic fusion of RGB and thermal features by mask-guided attention. These mentioned methods all assume that all RGB-T image pairs are well registered, and share the same bounding boxes. But the spatial shift of features between two modalities reduces the reliability of features within proposals after fusion. 

Some works have noticed this issue. MLPD~\cite{kim2021mlpd} takes the unpaired pedestrian between two modalities into consideration, which is caused by different FOVs of two cameras. In MBNet~\cite{mbnet2020}, pixel-level shifts in proposals are estimated densely, after which features will be aligned with deformable anchors. Kim \etal~\cite{kim2021uncertainty} consider not only low visibility but also incomplete pedestrian caused by shifts as the uncertainty of proposals, then descend the uncertainty by making proposals with higher uncertainty similar to those with lower uncertainty. Though the misregistration issue has been considered in ~\cite{kim2021uncertainty}, the modal-wise detection results are not given. Uncertainty-based methods~\cite{li2022confidence,kim2021uncertainty} have difficulties in modal-wise prediction, due to the degeneration of features with lower confidence during reweighting and representation, which is evaded in our work.

\subsection{Unregistered RGB-T Pedestrian Detection}
There are already some works that attempt to solve the unregistered issue explicitly and give modal-wise detection results at last. AR-CNN~\cite{zhang2019weakly} is the pioneering work of this task. As mentioned in Section I, modal-wise proposals are generated independently. Then the region feature alignment module is adopted to regress precise bounding boxes and shifts between modalities. Wan \etal~\cite{wanchaitanawong2021multi} propose a multi-modal RPN to get a pair of modal-wise proposals in the first stage of Faster-RCNN, then conduct a modal-wise box refinement in the second stage. Ni \etal~\cite{ni2022modality} use fusion features, which are composed of modal features weighted by illumination scores, to get a proposal, then conduct a modal-wise box refinement in the second stage. Yuan \etal~\cite{yuan2022translation} generalize AR-CNN and apply it in RGB-T vehicle detection, where reference modality is selected adaptively and original data augmentation named ``RoI jitter'' also includes more transformations.

Most of the methods for unregistered RGB-T pedestrian detection are based on Faster-RCNN and suffer from the incomplete information issue in proposals caused by misalignment, as  mentioned in Figure~\ref{Fig:intro}(a). Meanwhile, the confidence-based reweighting mechanism during feature fusion in these methods leads to the degeneration of low-score modality features. These two factors increase the error for modal-wise prediction. Additionally, all mentioned illumination-aware methods need an independently trained illumination subnetwork. Different from existing approaches, we propose an end-to-end dense detector without modal-wise reweighting, which resorts to a cross-modality proposal containing complete information for a pedestrian in each modality and predicts independent bounding boxes for both modalities.

\section{Proposed Method}
In this section, we present our proposed RGB-T pedestrian detector in detail. As shown in Figure~\ref{Fig:framework}, the proposed detector is composed of a two-stream backbone, the Path Aggregation Network (PAN)~\cite{pan} neck and Cross-Modality Proposal-guided
Feature Mining (CPFM) heads. The crux of the task is how to achieve inter-modality complementation with spatial misalignment. The two-stream backbone extracts the features from two modalities and fuses them together. With fusion features, the proposed CPFM head predicts the proposals enclosing two positions of a pedestrian in different modalities, avoiding the incompleteness of pedestrian information. Then the features characterizing the
pedestrians in a certain modality are mined adaptively to regress the precise bounding boxes in this modality. To deal with unpaired pedestrians, we introduce two kinds of NMS for post-processing.


\begin{figure}[!t]
\centering
\includegraphics[width=3.3in]{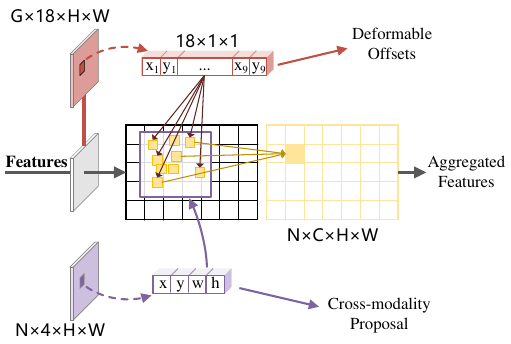}
\caption{Cross-modality proposal-guided feature mining. The cross-modality proposals (purple part) are predicted densely by the trunk branches in CPFM heads. Features inside proposals are mined adaptively with deformable convolutions. This figure takes $3\times3$ deformable convolution as an example, where $G$ is the number of groups of deformable offsets (red part). 
 }
\label{Fig:feature mining}
\end{figure}

\subsection{Two-stream Backbone}
The backbone aims to extract features from different modalities and obtain fusion features. We deploy two CSP-Darknet53s~\cite{yolov4} as two streams of the backbone (\ie~RGB and thermal), which take paired RGB and thermal images separately as input and output the concatenation-based fusion features. Some previous works\cite{cian,zhang2021guided} have indicated that the interaction between two modalities helps the modeling of modal-shared and modal-specific information. We follow this paradigm and deploy several interaction modules between modalities at each level. 

Especially, during the concatenation of features from unregistered RGB-T image pairs, features of a pedestrian from different streams could have a spatial shift. We deploy a spatial-aware interaction module to enhance the global reception, which is a transformer layer with co-attention modules (Co-T)~\cite{vilbert}.
The features from the \textit{k}-th stage in two streams $H_{rgb}^{(k)}$ and $H_{t}^{(k)},k\in\{3,4,5\}$ are fed into the Co-T module respectively, and the results are added back into each stream. Features from different streams are combined as an output $H_{out}^{(k)}$ by concatenation and convolution. The features after interaction are then fed into the next stage of each stream for further propagating.

Aside from $H_{out}^{(k)},k\in\{3,4,5\}$, as shown in Figure~\ref{Fig:framework}, the backbone outputs the stream-specific features $H_{rgb}^{(k)}$, $H_{t}^{(k)}$ for further use in the proposed CPFM head.

\subsection{Cross-modality Proposal-guided Feature Mining}\label{subB}
The proposed cross-modality proposal-guided feature mining aims to extract the valuable features for the pedestrian in each modality separately.
To achieve this goal, the proposals, which contain the complete information for a pedestrian in both modalities, are first predicted with fusion features. Then with the guidance of the proposals, features inside the proposals will be mined adaptively to find out the features characterizing targets in a certain modality.


\textbf{Cross-modality Proposal.}  The representation learned from fusion features helps to regress a bounding box that always contains two patterns of a pedestrian in different modalities, namely cross-modality proposals.
The trunk branch of CPFM head (purple part of the CPFM head in Figure~\ref{Fig:framework}) takes fusion features from PAN and predicts cross-modality proposals densely after four convolution layers and a $3\times3$ regression head. Aside from the regression head, a binary classification head, namely ``confidence head'' in object detection, is needed at the end of the branch to distinguish the foreground and background, which is not shown in Figure~\ref{Fig:framework} for concision. The proposals are supervised with the enclosing boxes of paired RGB-T ground truth boxes shown in Figure~\ref{Fig:intro}. Same as YOLOX~\cite{ge2021yolox}, each box is encoded by a 4D vector $(x', y', w', h')$ which means the normalized center location and size of the box relative to the anchor grid. The box can be decoded as:
\begin{equation}\label{ppsl_decode}
\begin{aligned}
x_{center}&=s_k \cdot x' + i \cdot s_k,\\
y_{center}&=s_k \cdot y' + j \cdot s_k,\\
w&=s_k \cdot e^{w'},\\
h&=s_k \cdot e^{h'},
\end{aligned}
\end{equation}
where $(i,j)$ is the location in feature map, and $s_k$ is the stride of backbone at \textit{k}-th stage. 

The cross-modality proposal encloses the information of a pedestrian from two modalities. With the guidance of proposals, we find out valuable features inside proposals for each modality adaptively by a local-attention mechanism, which is conducted with deformable convolution in practice. Instead of reusing fusion features from PAN, we add features from different branches of backbone $H_{rgb}^{(k)}$ and $H_{t}^{(k)},k\in\{3,4,5\}$ to enhance the modal-wise information.

\textbf{Feature mining.}  Feature mining will be conducted independently within the cross-modality proposals in different modalities to extract precise features, then followed by a regression head at last. Predictions of the proposal branch are first decoded to real scale at input image based on Eq.~\ref{ppsl_decode}, represented as $\mathbf{X}_0^{(k)} \in \mathbb{R}^{N\times 4 \times H \times W}$, where $N$ is the batch size, and $H \times W$ is the size of the feature map. Then upper and lower branches in the CPFM head deal with two modalities respectively. In the upper branch, for example, as shown in Figure~\ref{Fig:feature mining}, enhanced features with $C$ channels $\mathbf{F}_{rgb}^{(k)} \in \mathbb{R}^{N\times C \times H \times W}$ are fed into a convolution layer to predict the $3\times3$ deformable offsets $\mathbf{O}_{rgb}^{(k)} \in \mathbb{R}^{G\times (3\times3\times2) \times H \times W}$, which are then mapped into $[0,1]$ by a Sigmoid function. 

We define the center of deformable convolution as:
\begin{equation}\label{c_x}
\begin{aligned}
x_{ij}^c &= 0.5 + i,\\
y_{ij}^c &= 0.5 + j,
\end{aligned}
\end{equation}
where $(i,j)$ is the location in feature map. An element of $\mathbf{O}_{rgb}^{(k)}$ at location $(i,j)$, $o_{ij}$, is an encoded offset of $3\times3$ deformable convolution inside the proposal predicted at $(i,j)$. The final offset of deformable convolution at $(i,j)$ is an integration of the offset inside the proposal and the offset of the proposal itself. Let $x_{ij}=(x, y, w, h)$ be an element of $\mathbf{X}_0^{(k)}$ at location $(i,j)$, elements of the 4D vector mean top-left corner(need a transformation) and size of the predicted proposals in images. The decoding procedure can be formulated as:
\begin{equation}\label{o_decode}
\begin{aligned}
\hat{o}_{ij}^{(2n)} &= o_{ij}^{(2n)} \cdot \frac{w}{s_k} + (\frac{x}{s_k} - x_{ij}^c) - q_x,\\
\hat{o}_{ij}^{(2n+1)} &= o_{ij}^{(2n+1)} \cdot \frac{h}{s_k} + (\frac{y}{s_k} - y_{ij}^c) - q_y,
n \in \mathbb{Z},
\end{aligned}
\end{equation}
where $2n$ and $2n+1$ mean the even and odd terms respectively. Note that, $q_x$ and $q_y$ in Eq.~\ref{o_decode} is a constant vector indicating the primary offset for each location in a classic $3\times3$ convolution. In practice, all these procedures are conducted with matrix operation, getting a decoded $\mathbf{\hat{O}}_{rgb}^{(k)}$. Feature mining is then conducted by a deformable convolution layer (DCN): 
\begin{equation}
\mathbf{\hat{F}}_{rgb}^{(k)}=\Psi_{DCN}(\mathbf{F}_{rgb}^{(k)},\mathbf{\hat{O}}_{rgb}^{(k)}).
\end{equation}
The example of cross-modality proposal guided feature mining  with $3\times3$ deformable convolution is shown in Figure~\ref{Fig:feature mining}. The result of deformable convolution will be accumulated in the same location as the output features. Besides $3\times3$, the kernel size of deformable convolution can be larger for fine-grained feature mining, which will be introduced in Section \ref{sec4}.

\begin{figure}[!t]
\centering
\includegraphics[width=2.3in]{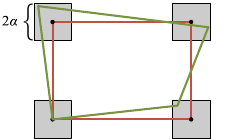}
\caption{Homography augmentation. Red and green boxes mean the RGB and thermal images respectively. Each corner of the thermal image is added with a random shift. The before- and after-moving corner pairs are used to solve a homography matrix to warp the thermal image.}
\label{Fig:homo}
\end{figure}

Same as the trunk branch, the upper and lower branch end with a regression head and a confidence head. The result of upper regression head $\mathbf{\Delta}_{rgb}^{(k)} \in \mathbb{R}^{N \times 4 \times H \times W}$ is the residual for cross-modality proposals $\mathbf{X}_0^{(k)}$ in box refinement. Let $d_{ij}=(d_1,d_2,d_3,d_4)$ be an element of $\mathbf{\Delta}_{rgb}^{(k)}$, the top-left and right-bottom corners of final bounding boxes in RGB image can be decoded with $d_{ij}$ and $x_{ij}$. The procedure is formulated as:
\begin{equation}\label{eq:delta}
\begin{aligned}
x_{tl}&=x+d_1 \cdot w,\\
y_{tl}&=y+d_2 \cdot h,\\
x_{rb}&=(x+w)+d_3 \cdot w,\\
y_{rb}&=(y+h)+d_4 \cdot h.
\end{aligned}
\end{equation}

\subsection{Homography Augmentation}
Inspired by AR-CNN~\cite{zhang2019weakly}, we design a data augmentation for our RGB-T detector, which expands the training set by simulating the misalignment between two images. Instead of modeling the transformation as a shift in AR-CNN, we consider the parallax and the difference in FOVs between two cameras, and generalize the shift transformation to a homography transformation.
The homography transformation considers not only the shifts between pedestrians but also the variance of scales in different modalities and the warp of projection in different views. 

Different from the jitter strategy in AR-CNN, which is conducted on the proposal level, our homography transformation warps the input image directly. Besides pedestrians, the warp of images also takes the background area without pedestrians into consideration, which improves the robustness of the network. As shown in Figure~\ref{Fig:homo}, each corner of the thermal image is added with a random shift $(t_x,t_y)$, which can be formulated as:
\begin{equation}
(t_x,t_y) \sim U(-\alpha,\alpha;-\alpha,\alpha),
\end{equation}
where $U$ means the uniform distribution.
The four before- and after-moving point pairs are used to solve the homography matrix $\mathbf{\Lambda} \in \mathbb{R}^{3\times3}$. 
All pixels of the input thermal image are then mapped to new positions according to $\mathbf{\Lambda}$. The ground truth bounding boxes defined with corner coordinates are also transformed by $\mathbf{\Lambda}$.

\subsection{Decoupled NMS and Pair-wise NMS}\label{subD}
 Non-maximum suppression (NMS) is necessary for our dense pedestrian detector. And precise bounding boxes must be given in both two modalities for unregistered RGB-T pedestrian detection. To this end, we investigate two NMS strategies for our two-branch box predictions. 
 
 \textbf{Decoupled NMS.}  For the predicted bounding boxes of two modalities in our CPFM head, we can conduct the traditional NMS separately in two modalities. Anchors close to pedestrians tend to get a higher quality in prediction. Thus the decoupled NMS selects the best bounding box for a pedestrian in each modality. This tends to bring a better score in evaluation where different modalities are tested separately. The disadvantage, however, is that bounding boxes of a specific pedestrian in different modalities could be kept at different anchors, which will need extra post-processing to obtain the correspondence between results in different modalities.

\begin{algorithm}[!t]
\caption{Pair-wise Non-Maximum Suppression}\label{alg:alg1}
\begin{algorithmic}
\STATE 
\STATE \textbf{Input}: set $\mathcal{P}$: all predictions from 3 scales, whose element is $p_{i}=(\mathbf{x}_{rgb}, \mu_{rgb}; \mathbf{x}_{t}, \mu_{t}) \in \mathcal{P}$. Threshold $\tau$.
\STATE \textbf{Initialize}: set $\mathcal{R}=\{\}$: the final result of detection.
\STATE \textbf{Run}:
\STATE \hspace{0.5cm} Get $\mu_g=max(\mu_{rgb},\mu_{t})$ for all $p_i \in \mathcal{P}$
\STATE \hspace{0.5cm} Remove the elements $\mu_g<\tau$, update $\mathcal{P}$
\STATE \hspace{0.5cm} Get $\hat{p}_i=(\mathbf{x}_g, \mu_g)$ by Eq.~\ref{bbox}, denote as $\mathcal{\hat{P}}$
\STATE \hspace{0.5cm} $\mathcal{\hat{I}}=\mathbf{NMS}(\mathcal{\hat{P}})$ \quad \% $\mathcal{\hat{I}}$ are the indexes of kept detections
\STATE \hspace{0.5cm} $\mathcal{R}=\mathcal{P}[\mathcal{\hat{I}}]$
\STATE \hspace{0.5cm} \textbf{for} $r_i \in \mathcal{R}$:
\STATE \hspace{1.0cm} Remove $(\mathbf{x}_m$, $\mu_m)$ in $r_i$, $m\in \{rgb,t\}$, if $\mu_m<\tau$
\STATE \hspace{0.5cm} \textbf{end}
\STATE \hspace{0.5cm} \textbf{return} set $\mathcal{R}$
\end{algorithmic}
\label{alg1}
\end{algorithm}
 
 \textbf{Pair-wise NMS.}  Considering the correspondence between detection results in two modalities after feature mining, the cross-modality proposal and boxes in two modalities of a specific person will be predicted by the same anchor (refers to ``anchor point'' or ``grids'' in anchor-free detectors) in different branches of CPFM head. So in NMS, we combine the predicted bounding boxes in the upper and lower branches as a group, which will be suppressed or kept together. Meanwhile, there exist unpaired pedestrians, especially in image pairs of different FOVs. We need to decouple two results in different modalities and suppress the low-score one in these cases. 
 
 In pair-wise NMS, the higher score of two boxes $\mathbf{x}_{rgb}$ and $\mathbf{x}_{t}$ will be considered as the score of the group $\mu_g$. 
And the bounding box of the group is an integration of two boxes based on their scores:
\begin{equation}\label{bbox}
\mathbf{x}_{g}=\alpha \mathbf{x}_{rgb} +\beta \mathbf{x}_{t},
\end{equation}
where $\alpha$ and $\beta$ are normalized weights. If the score of a box is lower than a threshold $\tau$, the box of the group will completely depend on the other one. This procedure can be formulated as:
\begin{equation}
\begin{aligned}
\alpha &= \frac{f(\mu_{rgb})}{f(\mu_{rgb})+f(\mu_{t})},\\
\beta &= \frac{f(\mu_{t})}{f(\mu_{rgb})+f(\mu_{t})},
\end{aligned}
\end{equation}
where $\mu$ means the score of each box and $f(\cdot)$ is a gate function defined as:
\begin{equation}
f(x)=
\begin{cases}
0 &x < \tau\\
x &x \ge \tau,
\end{cases}
\end{equation}
The procedure of the proposed pair-wise NMS can be depicted as Algorithm \ref{alg:alg1}.

The advantage of the proposed pair-wise NMS is that two bounding boxes in each element $r_i \in \mathcal{R}$ have the same identity. Despite that detection result in one modality would be suppressed due to the unpaired pedestrian issue, the remainder of $r_i$ still keeps the identity. 

\subsection{End-to-end learning}\label{subE}
\textbf{Decoupled head.}  Decoupled head is widely used in object detection~\cite{facalloss, ge2021yolox}, which divides localization and classification tasks into two branches without sharing weights. Aiming to facilitate the convergence of the network, we deploy an additional classification branch hidden in Figure~\ref{Fig:framework}, ending with a single-class classification head. Especially, the proposal branch has no classification head at the end of the classification branch, which provides only features for RGB and thermal classification branches.

\begin{figure*}[ht]
    \centering
    \includegraphics[width=0.47\linewidth]{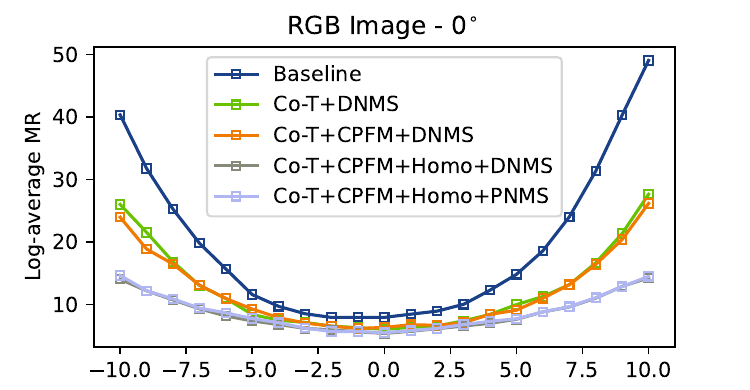}
    \includegraphics[width=0.47\linewidth]{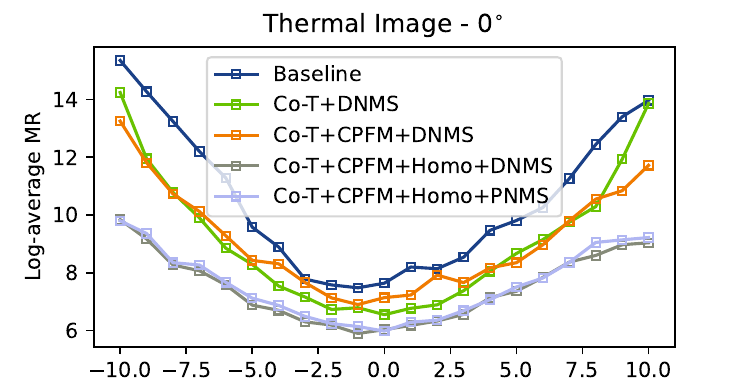}
    
    \includegraphics[width=0.47\linewidth]{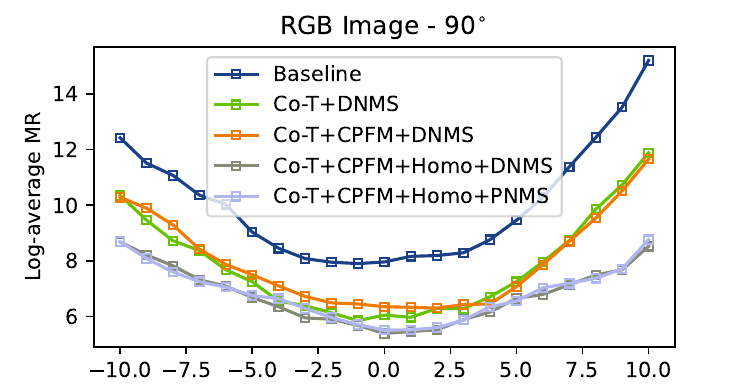}
    \includegraphics[width=0.47\linewidth]{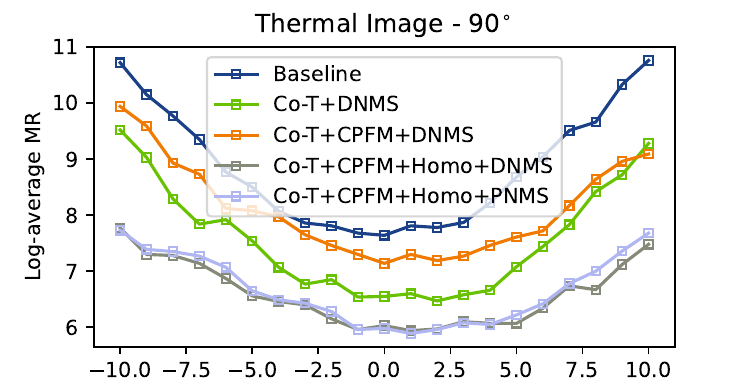}
    
    \includegraphics[width=0.47\linewidth]{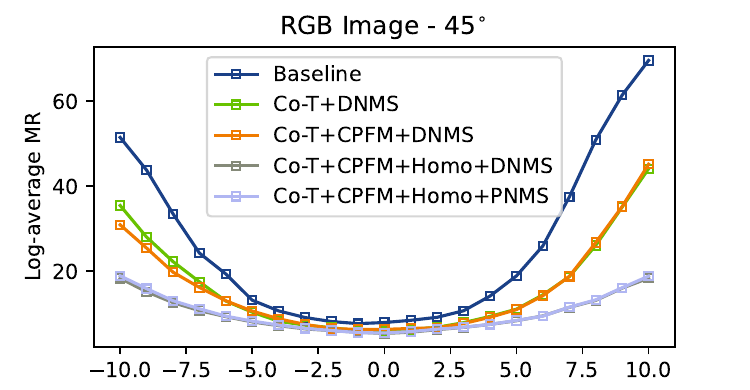}
    \includegraphics[width=0.47\linewidth]{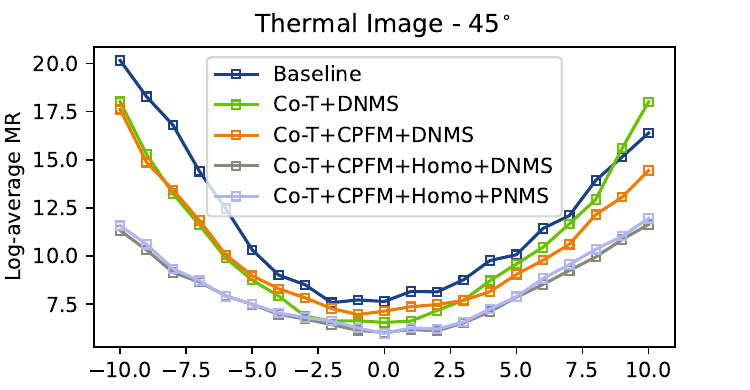}
    
    \includegraphics[width=0.47\linewidth]{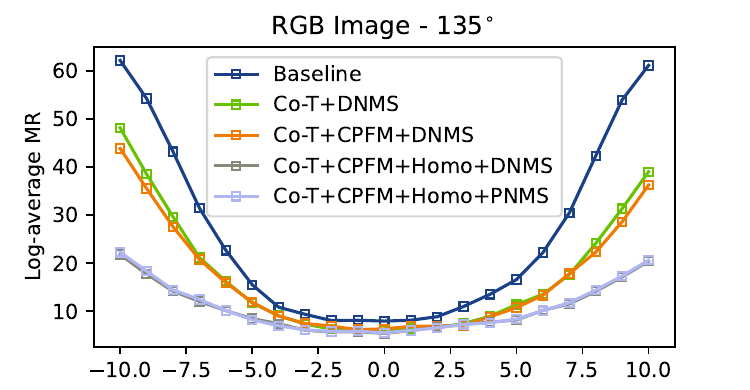}
    \includegraphics[width=0.47\linewidth]{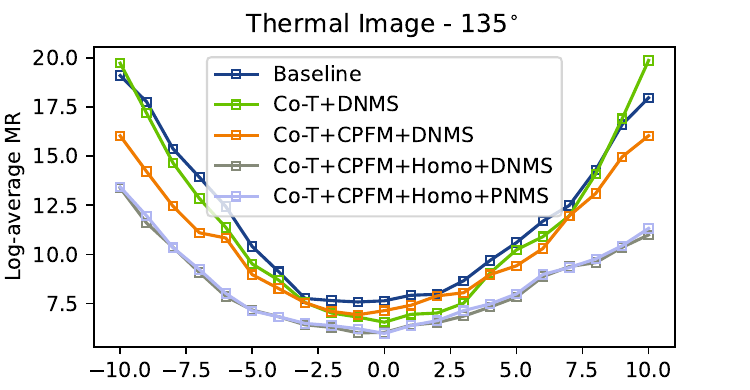}     
    
    \caption{Ablation study. We add spatial shifts manually between modalities to evaluate how each module helps the detection. All horizontal axes represent the degree of shift in pixels, denoted as $\delta_{shift}$. For the unregistered RGB-T pedestrian detection, we test the detection results of two modalities respectively with annotations on different modalities, \ie~left and right columns. Four rows show the results tested in different shift directions. }
    \label{fig:ablation}
\end{figure*}

\textbf{Total loss.}  The loss for training follows the conventional setting in dense object detection, which contains $L_{reg}$, $L_{conf}$, and $L_{cls}$. The loss of proposals, RGB, and thermal detections are added together. The total loss of the proposed network is formulated as:
\begin{equation}\label{eq:total_loss}
\begin{aligned}
L_{total}=&L_{reg}^{0}+L_{conf}^{0}+L_{reg}^{rgb}+L_{conf}^{rgb}+L_{cls}^{rgb}+\\
&L_{reg}^{t}+L_{conf}^{t}+L_{cls}^{t},
\end{aligned}
\end{equation}
where $L^0$,$L^{rgb}$,$L^{t}$ mean the losses in three branches respectively in CPFM head. The regression loss $L_{reg}$ in our work is calculated by Distance  Intersection over Union (DIoU)~\cite{diou} loss. The confidence loss $L_{conf}$ and classification loss $L_{cls}$ are both calculated by Binary Cross Entropy (BCE) loss. Following the paradigm of YOLOX, we will add a $L1$ loss for regression heads in the last few epochs of training. Note that the outputs of regression heads in the upper and lower branches are residual for proposals. To this end, the ground truth of these two heads in getting L1 loss should be encoded according to the inverse Eq.~\ref{eq:delta}.

With $L_{total}$, all the parameters can be trained jointly. During the optimization of $L_{total}$, the gradients of some paths are detached for the robustness of training, as shown in Figure~\ref{Fig:framework}. We use SimOTA~\cite{ge2021yolox} as the label assignment strategy, which is conducted on the proposal branch in CPFM heads. To be specific, if an anchor in the proposal branch is assigned to predict the cross-modality proposal of a certain pedestrian, the anchors at the same position in the upper and lower branches
will be assigned to the same pedestrian.

\section{Experiments}\label{sec4}
\subsection{Datasets and Evaluation Settings}
\textbf{Datasets.}  The KAIST~\cite{hwang2015multispectral} benchmark is widely used in RGB-T pedestrian detection. It provides more than 7000 pairs of RGB-T images with $512\times640$ resolution for training and more than 2000 pairs for testing. Due to the poor quality of original annotations, we use the annotations relabeled by Liu \etal~\cite{liu2016multispectral} conventionally for a modal-unified test of our method, where a pedestrian has only one bounding box in two modalities, as many existing works~\cite{ni2022modality,mbnet2020,kim2021uncertainty}. Though most pedestrian instances in the KAIST benchmark are well aligned, the remainder are unconscionable to share the same boxes in different modalities. Zhang \etal~\cite{zhang2019weakly} relabel the pedestrians in different modalities separately, which will be mainly used in our experiments to evaluate our method for unregistered pedestrian detection.

CVC-14~\cite{cvc14} is also a widely accepted dataset for RGB-T pedestrian detection, where RGB images are converted into gray mode. The annotations of CVC-14 are also labeled separately. CVC-14 benchmark is more challenging than KAIST due to larger shifts between box pairs. Based on our statistics, the average shift between paired boxes in CVC-14 is 17.32 pixels while the average shift in KAIST is only 1.4 pixels. 

\textbf{Evaluation Settings.}  As many works about pedestrian detection~\cite{zhang2019weakly, kieu2020task, kim2021uncertainty}, we conventionally adopt the log-average Miss Rate (LAMR) as the evaluation metric, where the range of False Positive Per Image (FPPI) is set to $[10^{-2},10^0]$ in log-space and the IoU threshold in calculating miss rate is 0.5. Note that, LAMR will be denoted as MR in the remainder of this section for concision, where a lower score is better.

\begin{table*}[ht]
\caption{Comparison With Other Methods On KAIST Benchmark. $\mathrm{MR \dagger}$ means the results on classic modal-shared annotation, $\mathrm{MR_R}$ and $\mathrm{MR_T}$ are tested with modal-specific labeled annotation} \label{tab:sota}
\centering
\begin{tabular}{c|c|ccc|ccc|ccc}
\Xhline{1pt}
\multirow{2}{*}{Methods} & \multirow{2}{*}{Backbone} & \multicolumn{3}{c|}{$\mathrm{MR \dagger}$}  & \multicolumn{3}{c|}{$\mathrm{MR_R}$} & \multicolumn{3}{c}{$\mathrm{MR_T}$} \\ \Xcline{3-11}{0.4pt} 
 & & Day & Night & All & Day & Night & All & Day & Night & All \\ \Xhline{0.4pt}
 Halfway Fusion~\cite{liu2016multispectral} & VGG-16 & 24.88 & 26.59 & 25.75 & 24.29 & 26.12 & 25.10 
 & 25.20 & 24.90 & 25.51 \\
 Fusion RPN~\cite{fusionrpn} & VGG-16 & 19.55 & 22.12 & 20.67 & 19.69 & 21.83 & 20.52 & 21.08 & 20.88 & 21.43 \\ 
 Adapted Halfway Fusion~\cite{zhang2019weakly} & VGG-16 & 15.36 & 14.99 & 15.18 & 14.56 & 15.72 & 15.06 & 15.48 & 14.84 & 15.59 \\
 IATDNN+IAMSS~\cite{iatdnn} & VGG-16 & 14.67 & 15.72 & 14.95 & 14.82 & 15.87 & 15.14 & 15.02 & 15.20 & 15.08 \\
 CIAN~\cite{cian} & VGG-16 & 14.77 & 11.13 & 14.12 & 15.13 & 12.43 & 14.64 & 16.21 & 9.88 & 14.68 \\
 MSDS-RCNN~\cite{msds} & VGG-16 & 10.60 & 13.73 & 11.63 & 9.91 & 14.21 & 11.28 & 12.02 & 13.01 & 12.51 \\
 AR-CNN~\cite{zhang2019weakly} & VGG-16 & 9.94 & 8.38 & 9.34 & 8.45 & 9.16 & 8.86 & 9.08 & 7.04 & 8.26 \\
 UGCML~\cite{kim2021uncertainty} & ResNet-50 & 8.18 & 6.96 & 7.89 & - & - & - & - & - & - \\
 CMPD~\cite{li2022confidence} & ResNet-50 & 8.77 & 7.31 & 8.16 & - & - & - & - & - & - \\
 MLPD~\cite{kim2021mlpd} & ResNet-50 & 8.36 & 6.35 & 7.61 & - & - & - & - & - & - \\
 \Xhline{0.4pt}
 CPFM + Decoupled NMS (Ours) & VGG16 & 8.4 & 6.8 & 7.68 & 8.03 & 4.08 & 7.08 & 8.57 & 4.05 & 7.03 \\
 CPFM + Pair-wise NMS (Ours) & VGG16 & 7.75 & \textbf{5.2} & 6.84 & 8.05 & 4.85 & 7.15 & 8.78 & 4.14 & 7.34 \\
 CPFM + Decoupled NMS (Ours) & CSP-Darknet53 & 7.64 & 6.93 & 7.29 & \textbf{6.45} & \textbf{3.81} & \textbf{5.39} & 7.47 & \textbf{3.63} & 6.03\\
 CPFM + Pair-wise NMS (Ours) & CSP-Darknet53 & \textbf{7.09} & 5.61 & \textbf{6.62} & 6.53 & 3.91 & 5.51 & \textbf{7.37} & 3.68 & \textbf{5.98} \\
 \Xhline{1pt}
\end{tabular}
\end{table*}

\subsection{Implementation Details}
We implement our proposed network on MMdetection~\cite{mmdet} toolbox. We deploy dual CSP-Darknet53s and VGG16s as the backbone in our experiments, all of which are pretrained on COCO~\cite{COCO}.  The kernel size of deformable convolution in feature mining is set to $5\times5$, and offset groups are set to 4. The bounds of shifts in proposed Homography augmentation are set to $\alpha=10$.  Apart from Homography, we also adopted Mosaic, Mixup, Resize, and other conventional data augmentations in the training phase. We train our model for 14 epochs on KAIST and CVC-14 by SGD optimizer, whose learning rate is set to 0.02. Specially, we will remove Mosaic and Mixup from the augmentation pipeline, decay the learning rate with a 0.1 multiplier, and add the L1 loss at the last 2 epochs. Other unmentioned structures and training settings are the same as the original YOLOX~\cite{ge2021yolox}. We train our model for about 2.5 hours on two RTX 3090Ti GPUs.

Before training, we pair up annotations from two modalities relabeled by Zhang \etal~\cite{zhang2019weakly} with the Hungarian algorithm. Bounding boxes of a pedestrian are fed into the network as a group, which is essential in label assignment.

\begin{figure}[!t]
    \centering
    \includegraphics[width=3.45in]{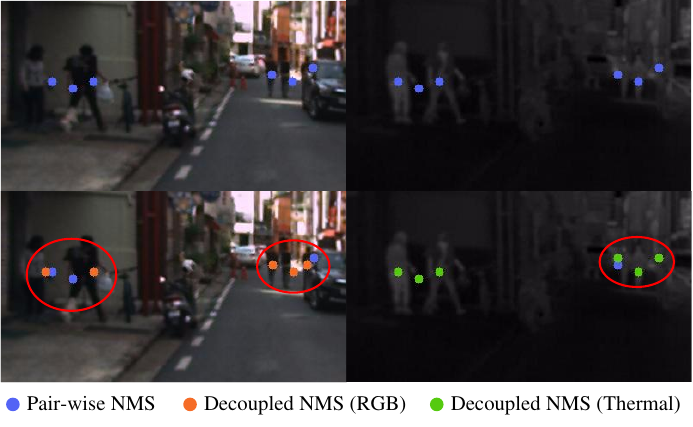}
    \caption{Visualization for pair-wise NMS (upper row) and decoupled NMS (lower row). The blue points are selected anchors after pair-wise NMS, which are shared in two modalities. The red and green points are selected anchors in two modalities after decoupled NMS respectively, some of which overlap with that in pair-wise NMS.}
    \label{fig:anchor_vis}
\end{figure}

\subsection{Ablation Study}\label{sec:ablation}
Our method is designed for unregistered pedestrian detection, components of which take the robustness to shift into consideration. We conduct further experiments in this section to find out how each component helps the robustness to shifts. Following the paradigm of AR-CNN~\cite{zhang2019weakly}, we add shifts between images from different modalities manually and test different combinations on different shifts. Same with that in AR-CNN, the directions of shifts are in $\{0^{\circ},90^{\circ},45^{\circ},135^{\circ}\}$ and shift levels $\delta_{shift}$ are set to $[-10,10]$ pixels. We conduct this on the KAIST benchmark. The final results are shown in Figure~\ref{fig:ablation}, where $0$ in x-axes means the original test set.

\vspace{1mm}
\noindent \textbf{Baseline.}  We implement a two-stream YOLOX-like detector as our baseline, which is a fully convolutional network and adopts decoupled NMS (DNMS) in different modalities. The training setting is the same as the original YOLOX. We test the RGB and thermal detection results with relabeled ground truth (GT) in \cite{zhang2019weakly} respectively. Log-average miss rates in the two modalities are 7.96 (RGB) and 7.64 (thermal). For comparison, we train the original YOLOX on two modalities independently, and get the results 18.64 and 16.02 respectively. The comparison shows the superiority of two-stream-based RGB-T detectors to those based on a single modality. The test results are shown as blue lines in Figure~\ref{fig:ablation}.

\vspace{1mm}
\noindent \textbf{Effect of Co-Attention Transformer Module.} 
Green curves show the MRs after using this module. Test results in the original test set (\ie~$\delta_{shift}=0$), where most pedestrians are well aligned, show the effectiveness of the modality interaction. And the global reception of co-attention transformer layers helps to
enlarge the scope of information integration, which contributes to the significant descent of MRs in large misalignment. 
However, the noise brought by mismatched information from the Co-T module strucks back to detection results when shifts become larger (\ie~$\delta_{shift} > 8$), as shown by green curves in thermal results in Figure~\ref{fig:ablation}.

\vspace{1mm}
\noindent \textbf{Effect of Cross-modality Proposal-guided Feature Mining.}  Proposed CPFM module supervised by enclosing boxes provides a prior scope for modal-wise feature mining. When $\delta_{shift}=0$, the proposed cross-modality proposal is equal to the traditional proposal in Faster-RCNN~\cite{fasterrcnn}. Compared with the local reception of $3\times3$ convolution layers, traditional proposal-based box refinement techniques enlarge the scope of information integration, which is important for large-scale objects and thus works well in the COCO dataset. Nevertheless, most of the pedestrians in KAIST are about 20 pixels in width. 
The area for information integration within the proposal in feature mining is similar to a $3\times3$ convolution. 
Feature mining in a small area is redundant and brings in noise. As shown in Figure~\ref{fig:ablation}, the MRs increase by about 0.3 and 0.6 respectively when $\delta_{shift}$ is equal to $0$.

Meanwhile, the area of a cross-modality proposal will become quite larger as the shift becomes larger. The information integrated within the cross-modality proposal is superior to that from a convolution layer. With the CPFM module (yellow curves), the performance is significantly improved in large shifts. Due to the inherent shift of the original test set, curves of RGB and thermal images are superior in opposite shift directions, as shown in Figure~\ref{fig:ablation}. Note that the results are aberrant when the shift direction is $90^{\circ}$, where misalignment exists along the longer axes of pedestrians. The normalized shifts are thus small, which brings in noise and is adverse to feature mining as mentioned above.

\vspace{1mm}
\noindent \textbf{Effect of Homography Data Augmentation.}  
Instead of being conducted on proposals, our proposed homography augmentation, following the paradigm of data augmentation for dense detectors, is conducted on the whole thermal images directly. It enriches the patterns of unregistered RGB-T image pairs and thus improves the performance of our RGB-T detector significantly in all shift directions for both modalities, as shown by gray curves in Figure~\ref{fig:ablation}. 

\begin{figure*}[!t]
    \centering
    \includegraphics[width=0.24\linewidth]{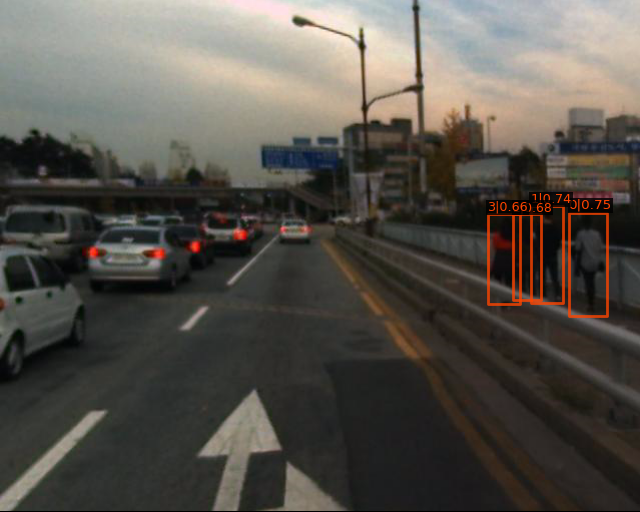}
    \includegraphics[width=0.24\linewidth]{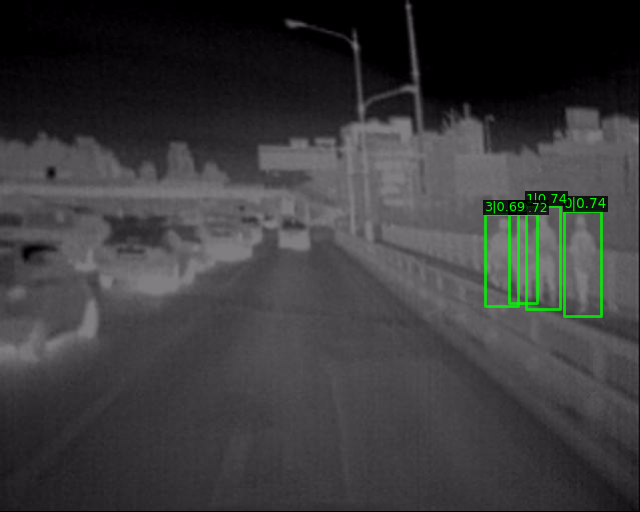}
    \includegraphics[width=0.24\linewidth]{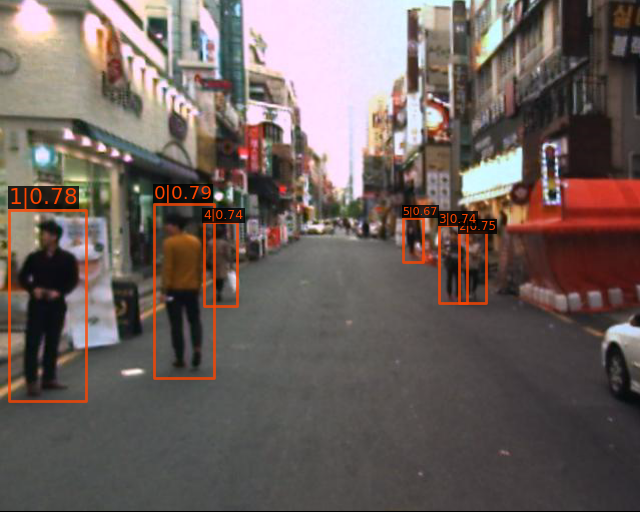}
    \includegraphics[width=0.24\linewidth]{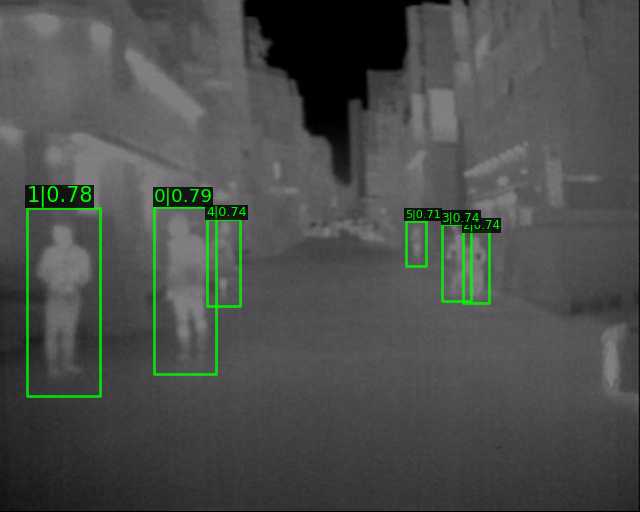}

    \vspace{0.05in}
    \includegraphics[width=0.24\linewidth]{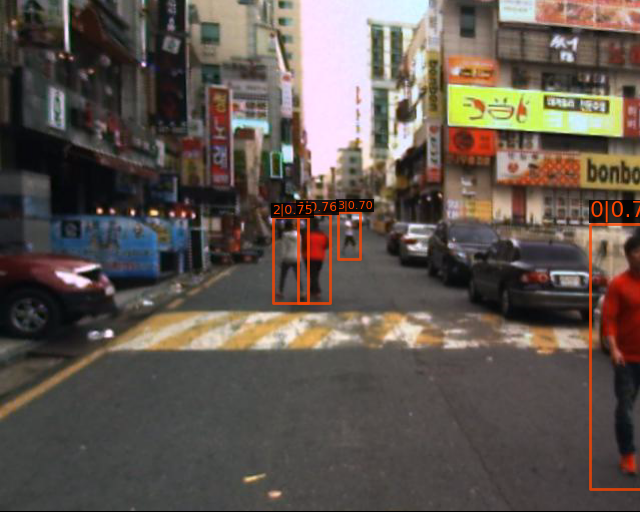}
    \includegraphics[width=0.24\linewidth]{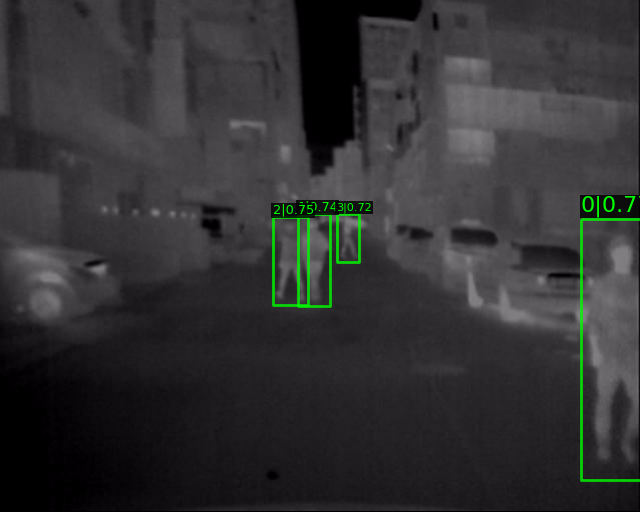}
    \includegraphics[width=0.24\linewidth]{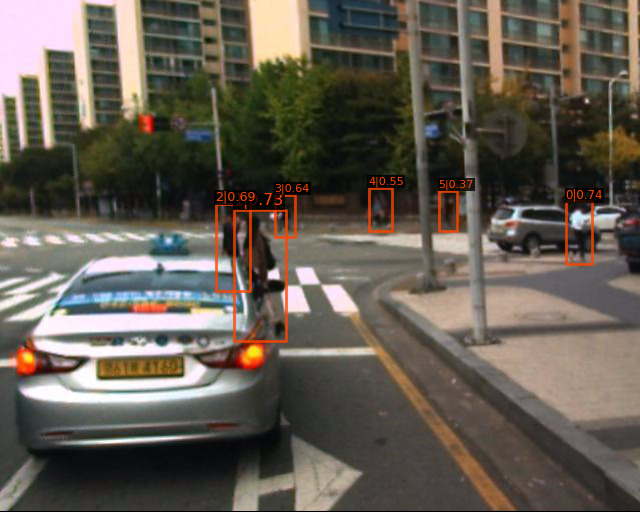}
    \includegraphics[width=0.24\linewidth]{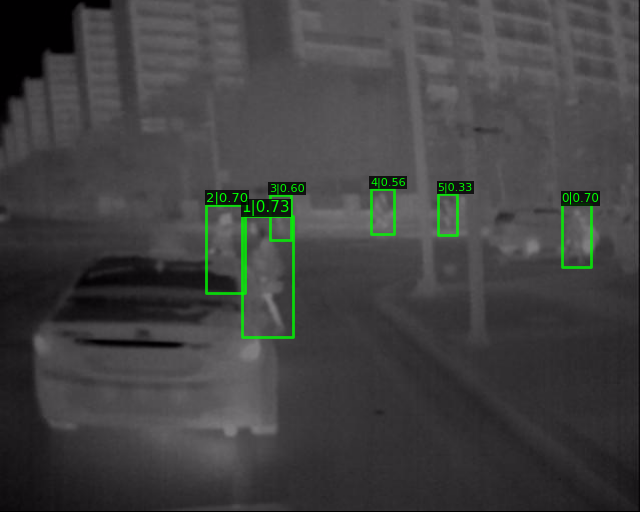}
    
    \caption{Typical detection results with pair-wise NMS in KAIST benchmark. The pedestrians locate at different positions in RGB and thermal images caused by the misalignment. Our method gives precise bounding boxes, which tightly enclose the pedestrian, for every pedestrian in both modalities. The score of each box is given independently. And every pedestrian is assigned an ID number shared by two boxes in different modalities.}
    \label{fig:demo}
\end{figure*}

\begin{table}[!t]
\caption{Comparison With Other Methods On CVC-14 Benchmark}\label{tab:cvc}
\centering
\begin{tabular}{c|c|ccc}
\Xhline{1pt}
\multirow{2}{*}{} & \multirow{2}{*}{Methods} & \multicolumn{3}{c}{MR} \\ 
\Xcline{3-5}{0.4pt}
                  &                          & Day   & Night   & All   \\
\Xhline{0.4pt}
\multirow{4}{*}{\rotatebox{90}{RGB}} & SVM~\cite{cvc14} & 37.6 & 76.9 & - \\
                  & ACF~\cite{park} & 65.0 & 83.2 & 71.3 \\
                  & Faster-RCNN~\cite{park} & 43.2 & 71.4 & 51.9 \\
                  & YOLOX~\cite{ge2021yolox} & 20.86& 26.63 & 23.29 \\
\Xhline{0.4pt}
\multirow{7}{*}{\rotatebox{90}{RGB-T}} & Halfway Fusion~\cite{park} & 38.1 & 34.4 & 37.0 \\
                  & Park \textit{et. al}~\cite{park} & 31.8 & 30.8 & 31.4 \\
                  & AR-CNN~\cite{zhang2019weakly} & 24.7 & 18.1 & 22.1 \\
                  & MLPD~\cite{kim2021mlpd} & 24.18 & 17.97 & 21.33 \\
                  & UGCML~\cite{kim2021uncertainty} & 23.87 & 11.08 & 18.7 \\
\Xcline{2-5}{0.4pt}
                  & CPFM + Decoupled NMS (Ours) & \textbf{20.63} & \textbf{9.45} & \textbf{15.47} \\
                  & CPFM + Pair-wise NMS (Ours) & 21.26 & 9.77 & 15.88 \\
\Xhline{1pt}
\end{tabular}
\end{table}

\vspace{1mm}
\noindent \textbf{Comparison for Pair-wise NMS and Decoupled NMS.}  According to the evaluation in Figure~\ref{fig:ablation}, two kinds of NMS strategies have comparable performance. In pair-wise NMS, a specific pedestrian in two modalities must be predicted with the same anchors. When the shift becomes larger, the pair-wise anchor for a pedestrian, selected after pair-wise NMS in some cases, is not optimal for both modalities in this pair. Considering that two modalities are tested independently in experiments, the decoupled NMS, which finds the optimal anchor for a pedestrian separately in different modalities, shows statistical superiority. As shown in Figure~\ref{fig:anchor_vis}, the first row shows the selected anchors after pair-wise NMS, where blue points are modal-sharing anchors and have the same location in two images. In decoupled NMS (second row), each pedestrian can choose the optimal anchor independently and eradicate the impact of the counterpart from the other modality.

\begin{figure*}[!t]
    \centering
    \includegraphics[width=0.24\linewidth]{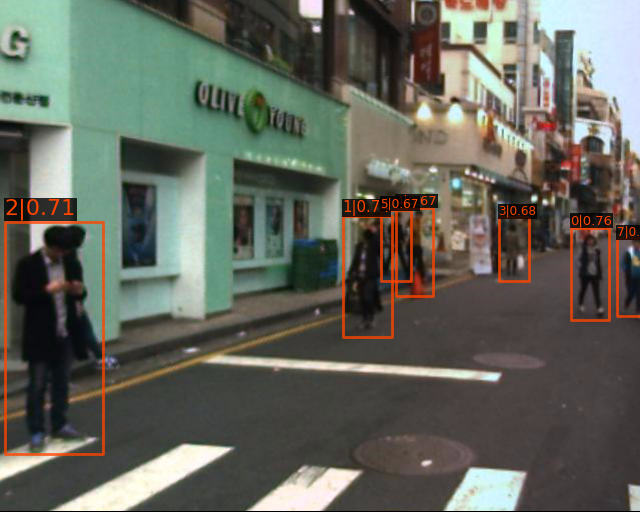}
    \includegraphics[width=0.24\linewidth]{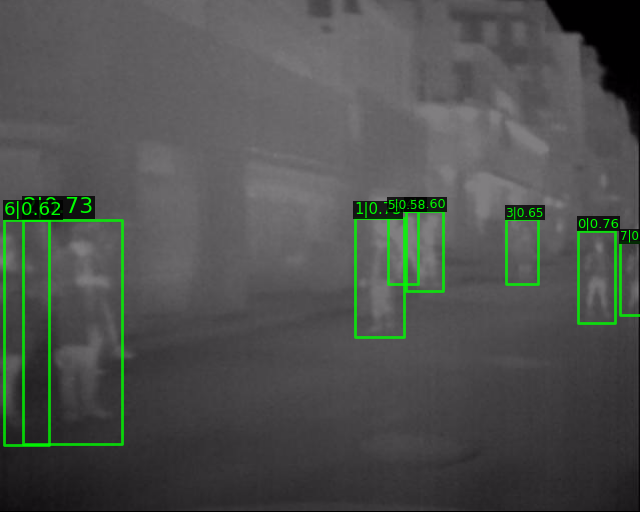}
    \includegraphics[width=0.24\linewidth]{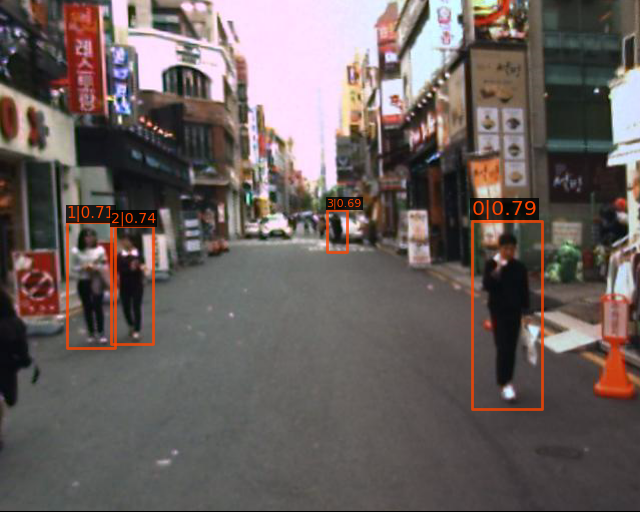}
    \includegraphics[width=0.24\linewidth]{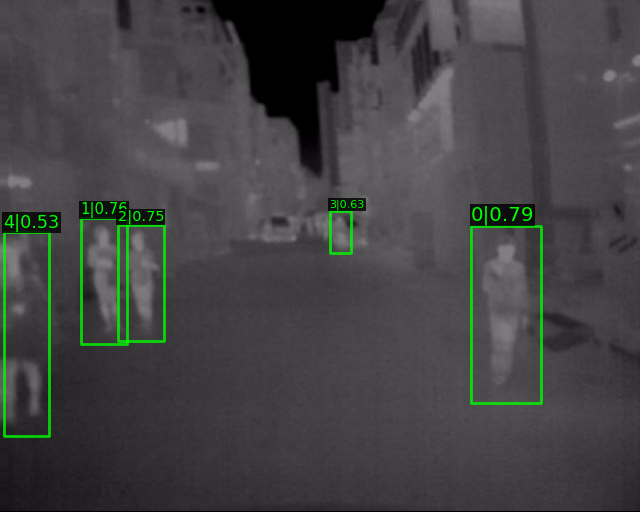}

    \vspace{0.05in}
    \includegraphics[width=0.24\linewidth]{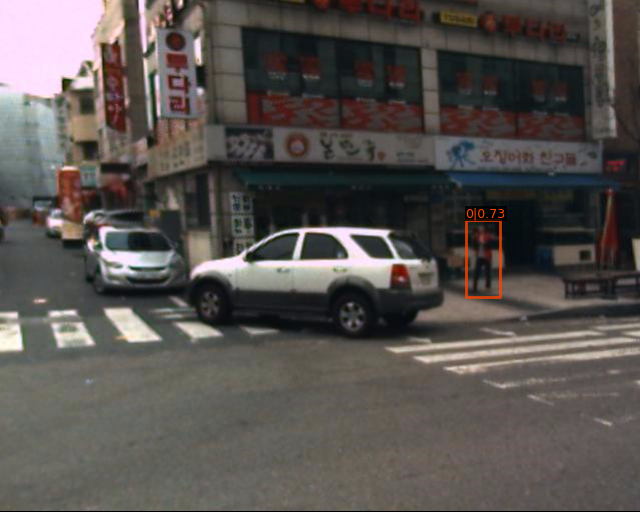}
    \includegraphics[width=0.24\linewidth]{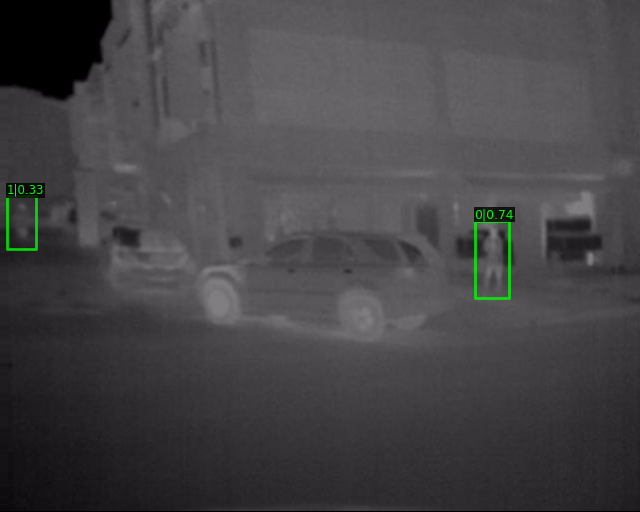}
    \includegraphics[width=0.24\linewidth]{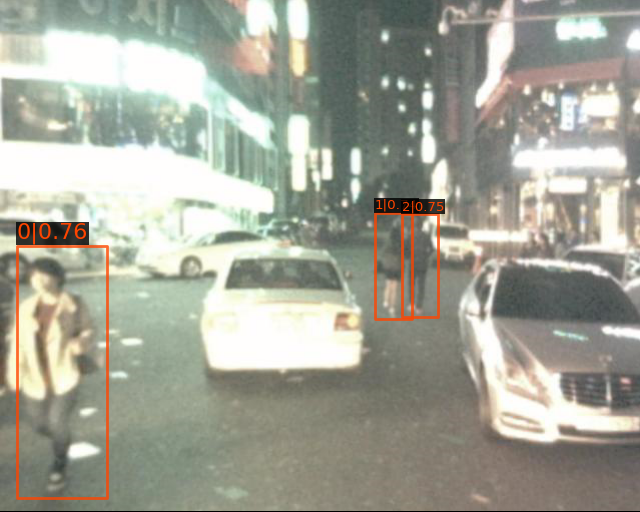}
    \includegraphics[width=0.24\linewidth]{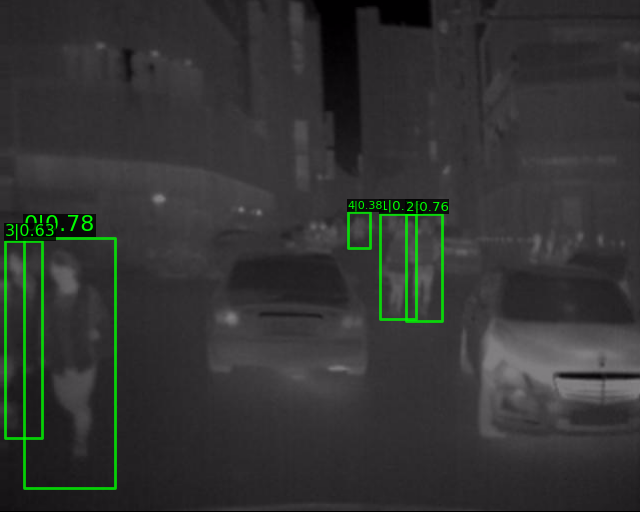}

    \caption{Unpaired cases in KAIST benchmark. The unpaired pedestrians at the corner are not only detected but also recognized. The recognized unpaired pedestrians are given boxes only in a single modality and assigned a new ID.}
    \label{fig:unpaired}
\end{figure*}

Meanwhile, the advantage of the proposed pair-wise NMS is that it keeps the correspondence in identity between detection results in different modalities. The pedestrian ID will be given directly without any other post-processing, which is ensured by the paired-up annotations in training. Some typical detection results with pair-wise NMS are shown in Figure~\ref{fig:demo} and Figure~\ref{fig:unpaired}. Figure~\ref{fig:demo} shows the effectiveness of our method in achieving the inter-modality complementation using spatially misaligned RGB and thermal pedestrian patches, which is one of the two challenges mentioned in Section~\ref{sec1}. And Figure~\ref{fig:unpaired} shows the unpaired cases, another challenge of our task. Our method assigns an independent ID number for the unpaired pedestrian and avoids the boxes for the background.

\subsection{Comparison with Other Methods}
 We make a comparison with other state-of-the-art methods for RGB-T pedestrian detection in both KAIST and CVC-14 benchmarks. Aside from the original test sets of two benchmarks, we utilize KAIST to test the robustness of our method to shifts between modalities.

\vspace{1mm}
\textbf{KAIST.}  As shown in Table~\ref{tab:sota}, we first conduct our comparison with the classic annotation, where a pedestrian in two modalities shares the same bounding boxes and the scores are marked with $\mathrm{MR \dagger}$. Then we utilize the KAIST-Paired annotation to conduct a further comparison in unregistered image pairs, where pedestrians have separate annotations for different modalities and scores are marked with $\mathrm{MR_R}$ and $\mathrm{MR_T}$ respectively. Following the definition of classic annotation, with pair-wise NMS, we obtain the enclosing boxes of paired detection boxes from two images as the final bounding boxes. For decoupled NMS, due to the lack of match in identity, we directly use the cross-modality proposals as the final boxes in the test for $\mathrm{MR \dagger}$. 

\begin{table*}[!t]
\caption{Robustness of Detectors to Different Shifts. O means the original MR without additional shifts. $\rho$ means the average MR at different shift levels}\label{tab:anti-shift}
\centering
\begin{tabular}{c|c|cccc}
\Xhline{1pt}
\multirow{2}{*}{Method} & \multirow{2}{*}{\makebox[0.05\linewidth][c]{O}} & \multicolumn{4}{c}{$\rho$} \\
\Xcline{3-6}{0.4pt}
                        &                    & \makebox[0.05\linewidth][c]{$0^{\circ}$} & \makebox[0.05\linewidth][c]{$45^{\circ}$} & \makebox[0.05\linewidth][c]{$90^{\circ}$} & \makebox[0.05\linewidth][c]{$135^{\circ}$} \\
\Xhline{0.4pt}
 Halfway Fusion~\cite{liu2016multispectral} & 25.51 & 33.73 & 36.25 & 28.3  & 36.71 \\
 Fusion RPN~\cite{fusionrpn}     & 21.43 & 30.12 & 31.69 & 24.48 & 34.02 \\
 CIAN~\cite{cian}           & 14.68 & 23.64 & 24.07 & 15.07 & 23.98 \\
 AR-CNN~\cite{zhang2019weakly}         & 8.26  & 9.34  & 9.73  & 8.91  & 9.79 \\
\Xhline{0.4pt}
 CPFM + Decoupled NMS (Ours) & 6.03 & \textbf{7.49} & \textbf{8.15} & \textbf{6.59} & \textbf{8.34} \\      
 CPFM + Pair-wise NMS (Ours) & \textbf{5.98} & 7.6 & 8.3 & 6.67 & 8.45 \\
\Xhline{1pt}
\end{tabular}
\end{table*}

For $\mathrm{MR \dagger}$, the results show that our method gets a significant superiority to the existing methods. With VGG-16, our method outperforms other methods with the same backbone by at least 26.8\% (with pair-wise NMS). Some recent methods adopt ResNet-50 as the backbone for better performance. Our method can even get a competitive performance with VGG-16 to those with ResNet-50. We also adopt CSP-Darknet53, which has a similar size to ResNet-50, in our backbone and get a superiority of 13\% to the SOTA method. With the pair-wise NMS, the enclosing box of two bounding boxes from different modalities can get a lower $\mathrm{MR \dagger}$, which owes to the proper correspondence between boxes in two modalities.

The proposed method can predict bounding boxes for two modalities separately. To this end, we test our method on KAIST-Paired annotation~\cite{zhang2019weakly} for a precise evaluation, where our method shows an even better performance at both day- and night-time. Most of the methods in Table~\ref{tab:sota} provide a more precise prediction for RGB images in the daytime but degenerate at night-time (shown as $\mathrm{MR_R}$), which indicates the superiority of different modalities. Different from these methods, our method performs better at night-time for both modalities (shown as $\mathrm{MR_R}$ and $\mathrm{MR_T}$). Considering that the detection results in RGB images are guided by the cross-modality proposals, this anomaly indicates the decisive role of thermal images in RGB-T pedestrian detection and the effectiveness of inter-modality complementation in our methods.

\textbf{CVC-14.}  Though the CVC-14 benchmark is annotated in both modalities, we just report the MRs in thermal images following the existing paradigm, which is shown in Table~\ref{tab:cvc}. Due to the larger misalignment and the gray mode of RGB images, CVC-14 is considered more challenging. Despite that all methods perform worse in CVC-14 than in KAIST, our method outperforms all the existing methods, which owes to the global reception of backbone and the effectiveness of the guidance of cross-modality proposals. Since the large misalignment in CVC-14, the decoupled NMS shows a better performance in statistical results.

\textbf{Robustness to Shifts.}  Shifts are added on thermal images, which is the same as that in section~\ref{sec:ablation}. As shown in Table~\ref{tab:anti-shift}, we choose several typical methods and report their performance in different shifts. We test these methods with shifts in four directions, whose levels are set to $[-10,10]$, and calculate the average of MRs in thermal images (marked with $\rho$). The O in Table~\ref{tab:anti-shift} means the original $\mathrm{MR_{T}}$ in Table~\ref{tab:sota}. When shifts become larger, decoupled NMS shows superiority. And our method shows a lower average MR than other methods in all directions, which means better robustness to shifts in all directions.

\section{Conclusion}
In this paper, we propose a dense RGB-T pedestrian detector for unregistered image pairs, which predicts the precise bounding boxes for pedestrians in both RGB and thermal images. To tackle the spatial misalignment of pedestrians in different modalities in feature fusion, we utilize the co-attention transformer between two streams to enhance the global reception. Then we conduct the feature mining with the guidance of cross-modality proposals, which are supervised with the enclosing boxes of ground truth boxes in two modalities, and regress the precise boxes in two modalities. We propose the homography data augmentation to simulate the conditions in unregistered image pairs, which provides a significant gain for the detector. At last, we investigate two NMS strategies, each of which has a unique advantage. The experiments show the effectiveness of our methods. In the future, we will study fine-grained identity matching for unregistered RGB-T pedestrians between modalities.

\bibliographystyle{IEEEtran}
\bibliography{ref.bib}
\end{document}